\documentclass[letterpaper]{article} 
\usepackage{aaai2026}  
\usepackage{times}  
\usepackage{helvet}  
\usepackage{courier}  
\usepackage[hyphens]{url}  
\usepackage{graphicx} 
\usepackage{natbib}  
\usepackage{caption} 
\usepackage{amsmath}
\usepackage{amssymb}
\usepackage{booktabs}
\usepackage{multirow}
\usepackage{colortbl}
\usepackage{algorithm}
\usepackage{algorithmic}

\usepackage{newfloat}
\usepackage{listings}
\newcommand{\modelname}{DEVIS-GRPO }

\DeclareCaptionStyle{ruled}{labelfont=normalfont,labelsep=colon,strut=off} 
\floatstyle{ruled}
\newfloat{listing}{tb}{lst}{}
\floatname{listing}{Listing}
\title{DEVIS-GRPO: Unleashing GRPO on Dynamic Extreme View Synthesis}
\author{
    Yi Zuo \textsuperscript{\rm 1,2}
    HuiMin Wu \textsuperscript{\rm 2},
    Lingling Li \textsuperscript{\rm 1},
    Fang Liu \textsuperscript{\rm 1},
    Licheng Jiao \textsuperscript{\rm 1},
    Qing Li \textsuperscript{\rm 2},
}
\affiliations{
    \textsuperscript{\rm 1}Xidian University\\
    \textsuperscript{\rm 2}State Key Laboratory of General Artificial Intelligence, BIGAI\\
}

\usepackage{bibentry}

\begin{document}

\maketitle

\begin{abstract}
Trajectory-controlled video generation has become essential for controllable video generation. While current methods perform well under small-view camera motions, they degrade significantly with large-view motions. Existing solutions for extreme-view synthesis typically require dedicated video pairs, demanding substantial annotation effort.
To address these limitations,
we propose \textbf{D}ynamic \textbf{E}xtreme \textbf{VI}ew \textbf{S}ynthesis-\textbf{GRPO} (\textbf{DEVIS-GRPO}),
a GRPO-based framework for trajectory-controlled video generation, the first online policy gradient method for extreme view video generation.
Central to our approach is a novel sampling strategy: \textbf{A}ccumulative \textbf{D}ynamic \textbf{E}xtreme \textbf{VI}ew \textbf{S}ynthesis (\textbf{ADEVIS}), which achieves large-view camera motions by progressively accumulating small-view increments. 
This method delivers two key advantages: 1) enhanced training efficiency, as it eliminates the need to warm-start the policy model by collecting expensive paired large-view videos, and 2) increased sampling diversity, achieved by flexibly varying trajectory configurations. 
Finally, we designed a multi-level consistency-quality reward function to select high-quality samples for model optimization.
Experiments on the Kubric-4D, iPhone, and DL3DV datasets demonstrate our method's superiority. On Kubric-4D, we achieve relative improvements of 21.57\% in PSNR and 7.31\% in SSIM over the second-best method in non-occlusion areas. On iPhone, LPIPS is reduced by 18.56\%.
\end{abstract}    
\section{Introduction}
\label{sec:intro}

\begin{figure*}
    \centering
    \includegraphics[width=1\textwidth]{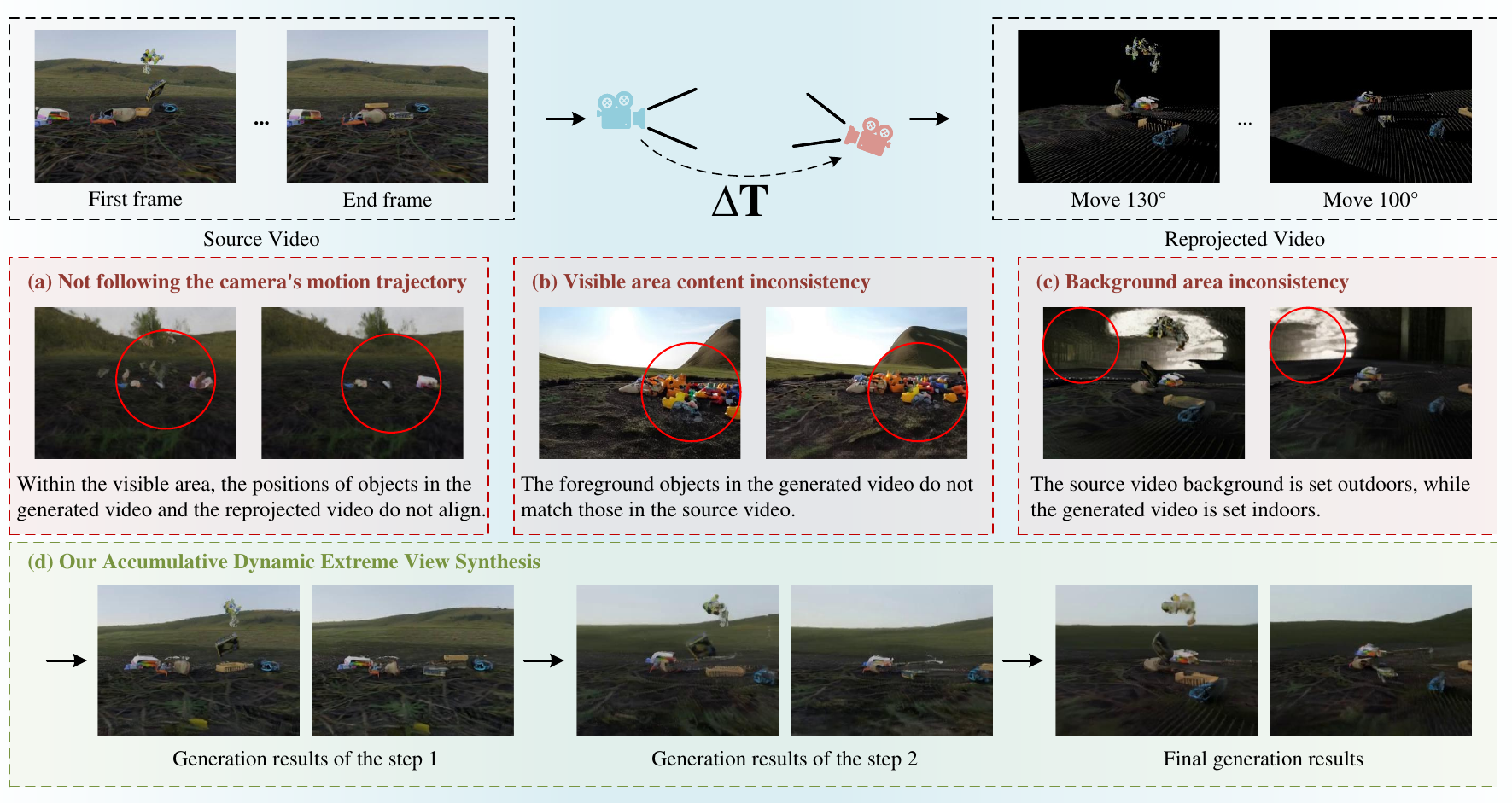}
    \caption{Under extreme viewpoints (large-view camera motions), existing methods suffer from two issues: (1) failure to follow the given camera trajectory; (2) inconsistency content between the generated video and the input video. Our ADEVIS addresses these problems by decomposing large-view camera motions into a series of small trajectory increments.}
    \label{fig:ExistedErr}
\end{figure*}
Trajectory-controlled video generation~\cite{he2024cameractrl,lee20253d,liu2025light,yu2024viewcrafter} aims to synthesize videos along specified camera trajectories, and has emerged as a fundamental problem in computer vision~\cite{liu2024sora, bar2024lumiere, chen2023videocrafter1}. This capability underpins diverse applications including virtual reality, autonomous driving simulation, and cinematic production~\cite{mei2022waymo, hu2023gaia, zuo2024edit}, where precise camera control is essential for creating immersive visual experiences.

Current approaches to trajectory-controlled video generation fall into two paradigms: explicit reprojection modeling~\cite{yu2025trajectorycrafter, hu2025ex} and implicit camera embedding modeling~\cite{bai2025recammaster, he2024cameractrl, van2024generative, kuang2024collaborative}. Both have achieved promising results in this domain.
However, existing methods struggle with large-view camera motions (extreme viewpoints). While they can successfully generate videos under small-view motions (typically within $40^\circ$), significant content inconsistency emerges when camera motions exceed $90^\circ$. 

As shown in Fig.~\ref{fig:ExistedErr} (a), given an extreme viewpoint, methods based on implicit camera embedding modeling cannot accurately follow the camera trajectory. In Fig.~\ref{fig:ExistedErr} (b) and (c), methods based on explicit reprojection modeling result in inconsistencies between the foreground and background content in the source video and the generated video.

This limitation stems from the scarcity of synchronized video datasets with extreme viewpoint changes. Existing datasets, constrained by real-world acquisition challenges~\cite{ren2022look}, contain only video pairs with small-angle camera motions or static scenes~\cite{zhou2018stereo,knapitsch2017tanks}, causing models to fail at generalizing to dynamic extreme viewpoints.

To address this challenge, we propose DEVIS-GRPO, a GRPO-based framework for trajectory-controlled video generation under extreme viewpoints. 
Our core insight is \textbf{A}ccumulative \textbf{D}ynamic \textbf{E}xtreme \textbf{VI}ew \textbf{S}ynthesis (\textbf{ADEVIS}), a novel sampling strategy. ADEVIS decomposes large-angle camera motions into smaller trajectory increments, then leverages prior knowledge from preceding steps to accumulately refine generated views (see Fig.~\ref{fig:ExistedErr} (d)), maintaining consistency during extreme viewpoint transitions. 

For DEVIS-GRPO, this sampling strategy offers two key advantages: (1) it eliminates the need to collect expensive paired large-view videos for warm-starting the policy model, and (2) it enhances sampling diversity by allowing flexible configuration of different accumulative step counts and parameters for camera motion at each step.
We then evaluate the videos generated by ADEVIS using a carefully designed multi-level consistency-quality reward function and optimize the model through group-based relative advantages.
During inference, we do not rely on multi-step accumulation; instead, we use an optimized model to generate large-angle videos directly in a single step. This ensures efficient inference.

Experiments demonstrate that DEVIS-GRPO outperforms existing methods in generating temporally and contextually coherent videos under extreme viewpoint. Ablation studies further validate the effectiveness of DEVIS-GRPO.
Our contributions are summarized as follows:

\begin{itemize}
    \item We first introduce GRPO to Dynamic Extreme View Synthesis and propose DEVIS-GRPO, which constructs diverse training groups by sampling videos from incremental trajectory steps and optimizes the model via designed multi-level consistency-quality reward function.
    \item We propose ADEVIS to provide paired training samples and diversified sampling strategies for DEVIS-GRPO, and design four accumulation strategies to explore this paradigm.
	\item Experiments on the Kubric-4D, iPhone and DL3DV datasets validate our method's superiority, significantly improving video consistency metrics, smoothness, and quality under extreme viewpoints.
\end{itemize}
\section{Related Work}
\label{sec:relatedwork}

\begin{figure*}
    \centering
    \includegraphics[width=1\textwidth]{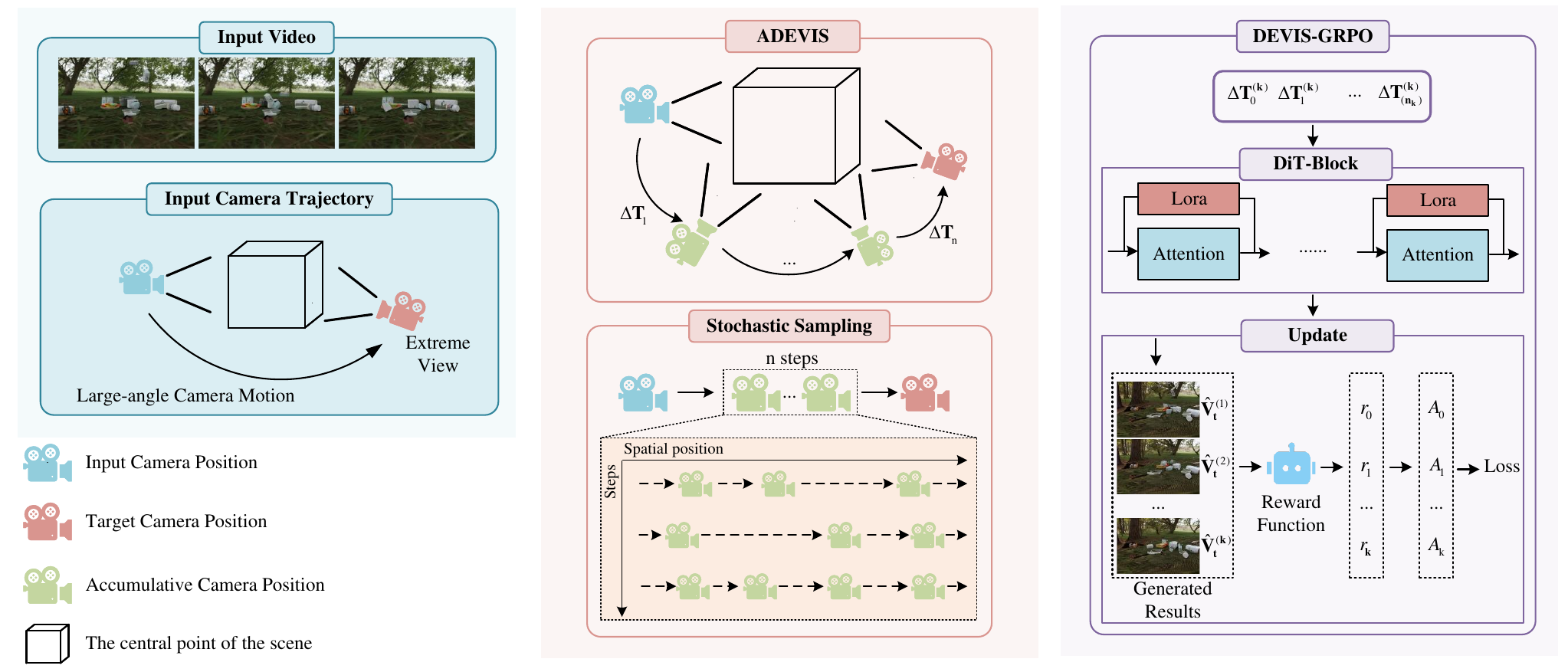}
    \caption{The pipeline of our proposed \modelname. We introduce Accumulative Dynamic Extreme View Synthesis (ADEVIS) to decompose large-angle camera motions into multiple small-angle motions (middle top). In DEVIS-GRPO, we randomize the number of sampling steps and camera motions (middle bottom), then design a reward function to score and compute relative advantages, finally optimizing the model via policy gradients (right).}
    \label{fig:img1}
\end{figure*}

\paragraph{3D and 4D Scene Reconstruction.}
Reconstruction-based approaches have achieved notable success in novel view synthesis. NeRF~\cite{mildenhall2021nerf} pioneered implicit neural representations, with subsequent works improving efficiency~\cite{muller2022instant} and anti-aliasing~\cite{barron2022mip}. 3D Gaussian Splatting (3DGS)~\cite{kerbl20233d} enables real-time explicit rendering with comparable quality. For dynamic scenes, 4D Gaussian Splatting methods~\cite{wu20244d, yang2024realtime, duan20244d} extend 3DGS with temporal modeling. Recent feed-forward approaches, including DUSt3R~\cite{wang2024dust3r} and MASt3R~\cite{leroy2024grounding}, enable sparse-view reconstruction without camera calibration, while MV-DUSt3R+~\cite{tang2025mv} further extends to efficient multi-view settings.
However, these reconstruction-based methods rely on observed content and struggle to hallucinate plausible details in occluded regions under extreme camera motions. Our framework addresses this by leveraging video generation models, enabling the synthesis of temporally coherent and visually consistent content for unseen regions.

\paragraph{Reinforcement Learning for Generative Models.}
Reinforcement learning has been widely adopted to align generative models with human preferences. RLHF~\cite{ouyang2022training} established this direction by fine-tuning language models with PPO~\cite{schulman2017proximal} based on human feedback. 
For visual generation, early methods such as DPOK~\cite{fan2023dpok} and DDPO~\cite{black2024training} demonstrated feasibility but faced scalability challenges. 
Recent advances including DPO~\cite{rafailov2023direct} and its visual adaptation Diffusion-DPO~\cite{wallace2024diffusion} simplify training by directly optimizing preference objectives. GRPO~\cite{guo2025deepseek} further improves efficiency through group-relative advantages, with DanceGRPO~\cite{xue2025dancegrpo} and Flow-GRPO~\cite{liu2025flowgrpo} extending it to image and video generation. However, these methods primarily target single image or video generation, without addressing content consistency across extreme viewpoints.
Our DEVIS-GRPO is the first to extend GRPO to camera-controlled extreme-view video generation, introducing stochastic trajectory grouping and multi-level consistency-quality reward function that evaluates consistency across viewpoints.

\paragraph{Camera-Controlled Video Generation.}

Recent video diffusion models~\cite{bar2024lumiere, blattmann2023stable, yang2025cogvideox} have enabled controllable video generation with various control signals~\cite{wang2024motionctrl, guo2023animatediff,qin2025worldsimbench}. 
Camera-controlled video generation~\cite{he2024cameractrl, bahmani2025vdd, zheng2024cami2v} specifically focuses on producing videos following specified camera trajectories. Based on how camera information is incorporated, existing methods can be categorized into two paradigms. 
\textit{Implicit camera embedding}, such as CameraCtrl~\cite{he2024cameractrl}, ReCamMaster~\cite{bai2025recammaster}, and GCD~\cite{van2024generative}, encode camera parameters as conditions for video diffusion models. While effective for small camera motions, these methods lack explicit geometric guidance for large viewpoint changes.
\textit{Explicit reprojection modeling}, including ViewCrafter~\cite{yu2024viewcrafter}, TrajectoryCrafter~\cite{yu2025trajectorycrafter}, and Follow-Your-Creation~\cite{ma2025follow}, leverage point cloud rendering or geometric warping to provide explicit 3D priors, then use diffusion models to complete occluded regions. 
However, due to the scarcity of training data and limited ability to handle large occlusions, these methods still struggle with extreme camera motions.
Our work addresses this limitation through ADEVIS, which decomposes extreme camera motions into sequences of manageable small-angle steps, combined with DEVIS-GRPO that enables effective training without synchronized multi-view supervision.

\section{Method}
\label{sec:method}

Our DEVIS-GRPO aims to generate a novel extreme viewpoint video $\mathbf{V_t}$ given a source video sequence $\mathbf{V_s}$ of a dynamic scene and a relative target camera trajectory $\Delta \mathbf{T}$ with camera direct motions.
Since existing methods perform poorly under large-angle camera motions, we propose an Accumulative Dynamic Extreme View Synthesis (ADEVIS) strategy to decompose extreme viewpoint transformations into a series of intermediate steps. We then designed DEVIS-GRPO to guide the baseline toward complex objectives.
Fig.~\ref{fig:img1} illustrates the pipeline of our DEVIS-GRPO.

\subsection{Preliminaries}
\label{sec_method:preliminaies}

\paragraph{The pipeline of TrajectoryCrafter.}
TrajectoryCrafter ~\cite{yu2025trajectorycrafter}, as one of the most advanced camera-controlled video generation methods, proposes a novel dual-stream conditional video diffusion model to integrate point cloud rendering with source video, thereby maintaining 4D consistency with the source video. 
First, the video is projected onto another viewpoint (after a camera motion of $\Delta \mathbf{T}$) via perspective reprojection:
\begin{equation}
    \mathbf{V'_t}, \mathbf{M'_t} = \Phi(\mathbf{V_s}, \Delta \mathbf{T}, \mathbf{D_s}),
\end{equation}
where $\Phi$ denotes the perspective reprojection, $\mathbf{D_s}$ is the depth maps of the source video $\mathbf{V_s}$, and $\mathbf{M'_t}$ is the binary mask indicating the occluded regions.
Then, the reprojected video $\mathbf{V'_t}$ is fed into a dual-stream conditional video diffusion model $\mathbf{\Theta_{diff}^{tc}}$ to generate the target video $\mathbf{V_t}$:
\begin{equation}
    \mathbf{\hat{V}_t} = \mathbf{\Theta_{diff}^{tc}}(\mathbf{V'_t}, \mathbf{M'_t}, \mathbf{V_s}, \mathbf{P_s}),
    \label{eq:gen}
\end{equation}
where $\mathbf{P_s}$ is the video caption of the source video $\mathbf{V_s}$ and $\mathbf{\hat{V}_t}$ is the generated video.
TrajectoryCrafter excels at handling small-angle changes, but it performs poorly at extreme viewpoints, especially in direct camera motion modes.

\paragraph{Group Relative Policy Optimization.} 
GRPO~\cite{guo2025deepseek} is a variant of PPO~\cite{schulman2017proximal} that does not require value model, instead estimating advantages based on group-level relative scores. For each query $q$ in the dataset, GRPO samples a set of outputs $\{o_1, o_2, \ldots, o_G\}$ from the old policy $\pi_{\theta_{\text{old}}}$, and then optimizes the policy model $\pi_{\theta}$ by maximizing the following objective:
\begin{align}
    \mathcal{J}_{\text{GRPO}}(\theta) &= \mathbb{E}_{q, \{o_i\}} \left[ \frac{1}{G}\sum_{i=1}^G \left( \mathcal{L}_{\text{clip}}^i - \beta D_{\text{KL}} \right) \right], \label{eq:grpo}\\
    \mathcal{L}_{\text{clip}}^i &= \min(pr_i A_i, \text{clip}(pr_i, 1-\epsilon, 1+\epsilon) A_i), \label{eq:clip}
\end{align}
where $pr_i = \pi_\theta(o_i|q) / \pi_{\theta_{\text{old}}}(o_i|q)$ is the probability ratio, and the KL divergence is:
\begin{equation}
    D_{\text{KL}} = \mathbb{D}_{\text{KL}}\left(\pi_{\theta} || \pi_{\text{old}}\right) = \frac{\pi_{\text{old}}(o_i|q)}{\pi_{\theta}(o_i|q)}- \log\frac{\pi_{\text{old}}(o_i|q)}{\pi_{\theta}(o_i|q)} - 1,
\end{equation}
where $\epsilon$, $\beta$ are hyperparameters, and $A_i$ is the advantage computed using the rewards within each group:
\begin{equation}
    A_i = \frac{r_i - \text{mean}(\{r_1, r_2, \ldots, r_G\})}{\text{std}(\{r_1, r_2, \ldots, r_G\})}.
    \label{eq:advantage}
\end{equation}

\label{sec:preliminary}

\begin{figure}[t]
    \centering
    \includegraphics[width=0.48\textwidth]{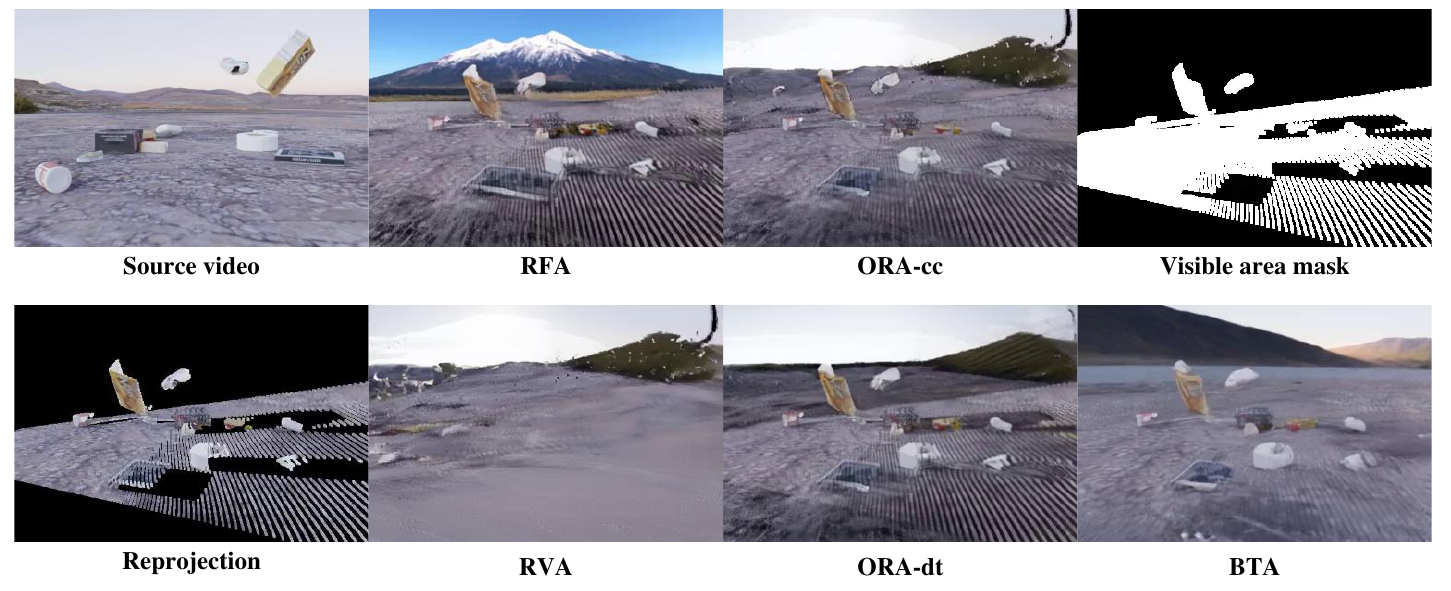}
    \caption{Visualization results of different accumulation methods. Rotate the viewpoint by more than 90 degrees.}
    \label{fig:img6}
\end{figure}

\begin{table*}[htbp]
  \centering
  \caption{Quantitative results in iphone and Kubric-4D Dataset with Large-angle Camera Motions. The best value is \textbf{bolded}.}
      \scalebox{0.9}{
    \begin{tabular}{c|c|ccccc|cccc}
    \toprule
    \multirow{2}[4]{*}{Dataset} & \multirow{2}[4]{*}{Method} & \multicolumn{5}{c|}{Consistency  Metrics} & \multicolumn{4}{c}{Video Smooth \& Quality} \\
\cmidrule{3-11}          &       & PSNR  & SSIM  & LPIPS & PSNR-Nocc & SSIM-Nocc & TF    & MS    & AQ    & IQ \\
    \midrule
    \multirow{4}[4]{*}{Kubric-4D} & GCD   & 13.50 & 0.3556 & 0.5631 & 13.78 & 0.8619 & 0.9648 & 0.9838 & 0.4080 & 0.4173 \\
          & EX-4D & 12.38 & 0.3384 & 0.6017 & 16.59 & 0.8875 & 0.3717 & 0.5145 & 0.3717 & 0.5145 \\
          & TrajectoryCrafter & 9.91  & 0.2728 & 0.6560 & 13.06 & 0.8417 & 0.9522 & 0.9773 & 0.3880 & 0.6000 \\
\cmidrule{2-11}          & Ours  & \textbf{13.59} & \textbf{0.436} & \textbf{0.5289} & \textbf{20.17} & \textbf{0.9524} & \textbf{0.9698} & \textbf{0.9867} & \textbf{0.4105} & 0.3828 \\
    \midrule
    \multirow{4}[4]{*}{Iphone} & GCD   & 11.44 & 0.1647 & 0.7708 & 10.39 & 0.5851 & 0.9744 & 0.9815 & 0.3511 & 0.4219 \\
          & EX-4D & 9.58  & 0.1668 & 0.6175 & 13.81 & 0.6168 & 0.9499 & 0.9645 & 0.3833 & 0.6996 \\
          & TrajectoryCrafter & 11.43 & 0.2081 & 0.5447 & 14.02 & \textbf{0.6543} & 0.9812 & 0.9802 & 0.4204 & 0.6118 \\
\cmidrule{2-11}          & Ours  & \textbf{13.63} & \textbf{0.2503} & \textbf{0.4436} & \textbf{15.05} & 0.6535 & \textbf{0.9843} & \textbf{0.9888} & \textbf{0.4219} & \textbf{0.7035} \\
    \midrule
    \multirow{3}[2]{*}{DL3DV} & Ex4D  & 10.30 & 0.1834 & 0.6071 & 10.97 & 0.4859 & 0.8855 & 0.9292 & 0.3486 & 0.4976 \\
          & TrajectoryCrafter & 11.05 & \textbf{0.2295} & 0.5690 & 11.24 & 0.5724 & 0.9262 & \textbf{0.9763} & 0.4460 & 0.5806 \\
          & Ours  & \textbf{11.50} & 0.2185 & \textbf{0.5568} & \textbf{11.95} & \textbf{0.6024} & \textbf{0.9338} & 0.9741 & \textbf{0.4621} & \textbf{0.5825} \\
    \bottomrule
    \end{tabular}%
    }
  \label{tab:quantitative}%
\end{table*}%

\begin{figure*}[t]
    \centering
    \includegraphics[width=1\textwidth]{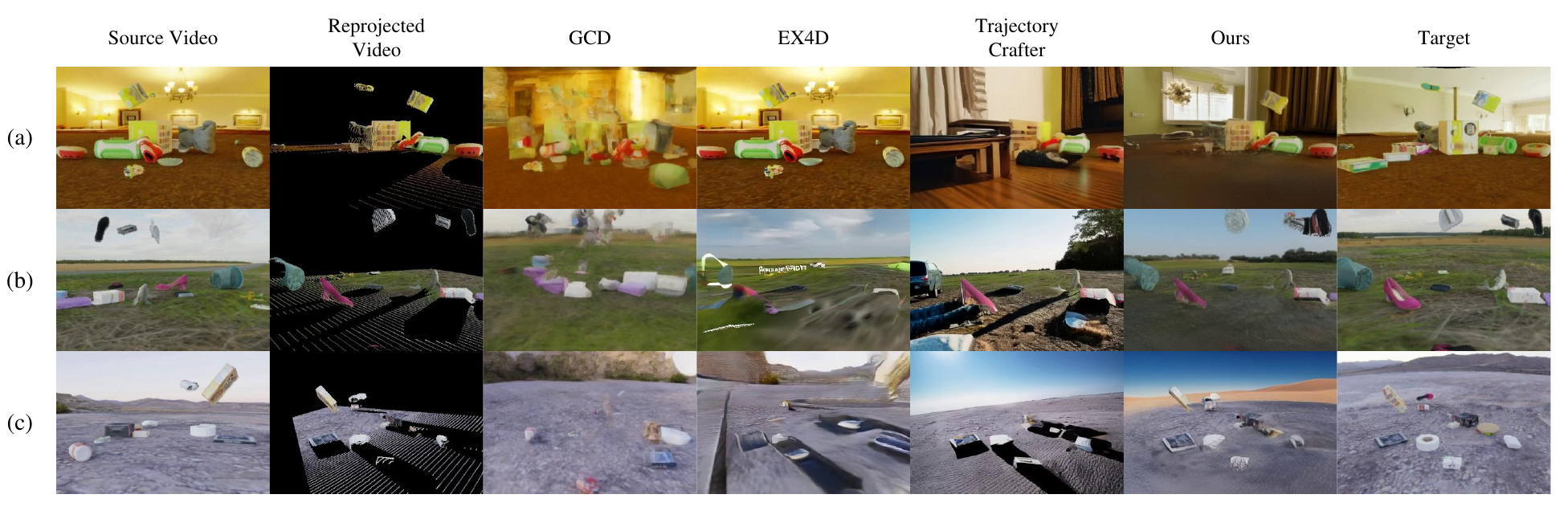}
    \caption{Qualitative comparison on Kubric-4D. Our proposed DEVIS-GRPO achieves consistency with both reprojected video and input video simultaneously across foreground and background regions, while the generated video fully adheres to the specified camera trajectory.}
    \label{fig:QualitativeKubric}
\end{figure*}

\subsection{Accumulative Dynamic Extreme View Synthesis}
\label{sec_method:adevis}

Existing baseline models are limited to generating videos with only small camera view changes. To obtain synchronized videos from extreme viewpoints, our key idea is to decompose large camera movements ($\Delta\mathbf{T}$) into a sequence of small view increments ($\Delta\mathbf{T}_i$), 
\begin{equation}
\Delta\mathbf{T}=\Delta\mathbf{T}_1+\Delta\mathbf{T}_2+\cdots+\Delta\mathbf{T_n},
\end{equation}
and then generate the video by progressively accumulating these increments. 
This leads to the reprojected video $\mathbf{V'_{i}}$ playing a crucial role in generating geometrically consistent video.

To obtain this geometry prior, we investigate four different strategies: Reference Video Accumulation, Reprojection Video Accumulation, Occlusion Region Accumulation, and Bullet Time Accumulation.

\paragraph{Reference Video Accumulation (RFA).}
In this variant, we use the generated video from the previous step as the reference video to replace $V_s$ in Eq.~\eqref{eq:gen}. 
The most straightforward way is to directly reproject from the source video $\mathbf{V'_i}$ at each accumulative step $i$,
which is defined as:
\begin{equation}
    \mathbf{V'_{rfa,i}}, \mathbf{M'_{rfa,i}} = \Phi(\mathbf{V_s}, \sum_{j=1}^{i} \Delta \mathbf{T}_j, \mathbf{D_s}).
\end{equation}
However, we empirically observe that this strategy produces large occluded regions due to the substantial view difference, which poses significant challenges for subsequent diffusion-based rendering.

\paragraph{Reprojection Video Accumulation (RVA).}
In this regard,
we explore an accumulative strategy by reprojecting from previously generated video $\mathbf{\hat{V}_{i-1}}$.
\begin{equation}
    \mathbf{V'_{rva,i}}, \mathbf{M'_{rva,i}} = \Phi(\mathbf{\hat{V}_{i-1}}, \Delta \mathbf{T}_i, \mathbf{\hat{D}_{i-1}}),
\end{equation}
where $\mathbf{\hat{D}_{i-1}}$ is the depth maps of $\mathbf{\hat{V}_{i-1}}$ obtained from the depth estimation model $\Phi_{depth}$.

\paragraph{Occlusion Region Accumulation (ORA).}
Although $\mathbf{V'_{rva, i}}$ can inherit all priors from the previous step, the error between the estimated depth $\Phi_{depth}$ and the true depth causes a deviation between the' reprojection of $\mathbf{V'_{rva, i}}$ and the true reprojection.
To address this issue, we designed ORA strategy, which fuses partial $\mathbf{V'_{rva,i}}$ and $\mathbf{V'_{i}}$ in the visible region.
Within $\mathbf{V'_{rva,i}}$, we consider the following two scenarios worthy of fusion: (1) Closer regions, as the error in reprojection increases with distance; (2) Continuous regions, where depth estimation models perform poorly in predicting depth along object boundaries in video.

We designate these as ORA-dt and ORA-cc, respectively. For ORA-dt, we set a threshold, treating areas below it as valuable. For ORA-cc, we employ the eight-connected domain algorithm to extract contiguous regions, then similarly apply a threshold, considering areas exceeding it as valuable.

\paragraph{Bullet Time Accumulation (BTA).}
In ORA-dt and ORA-cc, although key regions in $\mathbf{V'_{rva,i}}$ are filtered using depth thresholds and connected region thresholds, this strategy is not perfect. As shown in Fig.~\ref{fig:img6}, misalignments still occur at the junctions between $\mathbf{V'_{i}}$ and $\mathbf{V'_{rva,i}}$.

To avoid reprojection errors and preserve the prior information from the previous step, we propose BTA.
Specifically, we create a bullet time to seamlessly transition between the two viewpoints. We repeat the first frame of the source video $m$ times and decompose the corresponding $\Delta\mathbf{T}_i^1$ into smaller, consecutively increasing changes in viewpoint. The new source video and trajectory can be represented as:
\begin{equation}
    \mathbf{V_s}^{*} = \{\underbrace{\mathbf{f_1}, \mathbf{f_1}, \ldots, \mathbf{f_1}}_{\mathbf{m}\text{ times}}, \mathbf{f_2}, \ldots, \mathbf{f_{l}}\},
\end{equation}

\begin{equation}
    \Delta\mathbf{T_i^{*}} = \{\underbrace{\Delta\mathbf{t_i^1},\Delta\mathbf{t_i^2}, \ldots, \Delta\mathbf{t_i^{m}}}_{\mathbf{m} \text{ times}}, \Delta\mathbf{T_i^2}, \ldots, \Delta\mathbf{T_i^l}\},
\end{equation}
\begin{equation}
    \Delta\mathbf{t_i^j} = \frac{\Delta\mathbf{T_i^1} - \Delta\mathbf{T_{i-1}^1}}{\mathbf{m}} \times \mathbf{j}.
\end{equation}
where $\mathbf{l}$ is the length of $\mathbf{V_s}$. We then obtain the reprojected video $\mathbf{V_i}^{'}$ through perspective transformation $\Phi$,
\begin{equation}
    \mathbf{V_i}^{'} = \begin{cases}
        \Phi(\mathbf{V_s}^{*}, \Delta\mathbf{T_i^{*}}, \mathbf{D_s}) & \text{if } i = 1, \\
         \Gamma(\mathbf{\hat{V}_{i-1}}) \oplus \Phi(\mathbf{V_s}^{*}, \Delta\mathbf{T_i^{*}}, \mathbf{D_s}) & \text{if } i > 1,
    \end{cases}
\end{equation}
where $\Gamma(\mathbf{\hat{V}_{i-1}})$ is the bullet time part of $\mathbf{\hat{V}_{i-1}}$ and $\oplus$ denotes the concatenation operation on time dimension. 
After $n$ iterations and accumulations, we generate the final video $\mathbf{\hat{V}_n}$.

\subsection{Dynamic Extreme View Synthesis-GRPO}
\label{sec_method:devis-grpo}

Although ADEVIS can generate highly consistent extreme results, it consumes excessive inference time. To enable our model to generate videos under extreme viewpoints in a single step, we propose \textbf{D}ynamic \textbf{E}xtreme \textbf{VI}ew \textbf{S}ynthesis-\textbf{GRPO} (DEVIS-GRPO).
Specifically, we stochastically sample the number of accumulation steps $\mathbf{n}$ and the magnitude of camera motion at each step to create diverse trajectory groups. We then design a consistency-quality reward function to evaluate the quality of generated samples and optimize the model parameters via GRPO.

\paragraph{Stochastic Sampling}
We consider different trajectories sampled from the same $\Delta\mathbf{T}$ as a group.
For the $\mathbf{k}$-th sample, we randomize the number of accumulative steps $\mathbf{n_k}$ and the distance of motion in each step, where the motion distances within each sample are not necessarily equal:
\begin{equation}
    \Delta \mathbf{T_i^{(k)}} \neq \Delta \mathbf{T_{i+1}^{(k)}}, \quad \forall \mathbf{i} = 1, 2, \ldots, \mathbf{n_k}-1.
\end{equation}
We then generate the video $\mathbf{\hat{V}^{(k)}_t}$ using the ADEVIS with $\mathbf{n_k}$ steps.

\begin{figure*}[!t]
    \centering
    \includegraphics[width=0.95\textwidth]{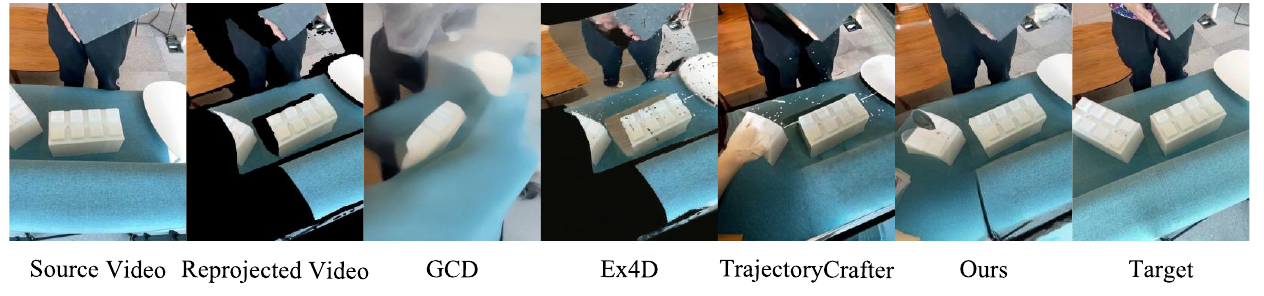}
    \caption{Qualitative comparison on Iphone. We compare DEVIS-GRPO with state-of-the-art methods GCD~\cite{van2024generative}, TrajectoryCrafter~\cite{yu2025trajectorycrafter}, and EX-4D~\cite{hu2025ex}.}
    \label{fig:QualitativeIphone}
\end{figure*}
\paragraph{Consistency-Quality Reward.} 
In multi-view video generation, maintaining consistency between the generated video and the source video is crucial. Therefore, we evaluate the quality of the generated video $\mathbf{\hat{V}_{t}^{(k)}}$ based on both pixel-level, semantic-level fidelity and video quality.

For pixel-level fidelity, we calculate the PSNR and SSIM between the reprojected source video $\mathbf{V_t^{'(k)}}$ and the generated video $\mathbf{\hat{V}_{t}^{(k)}}$ in the visible regions, expressed as:
\begin{equation}
    p_j = \mathcal{P}(\mathbf{V_t^{'(k)}} \odot \mathbf{M}, \mathbf{\hat{V}_{t}^{(k)}} \odot \mathbf{M}),
\end{equation}
\begin{equation}
    s_j = \mathcal{S}(\mathbf{V_t^{'(k)}} \odot \mathbf{M}, \mathbf{\hat{V}_{t}^{(k)}} \odot \mathbf{M}),
\end{equation}
where $\mathcal{P}(\cdot)$ denotes the PSNR metric, $\mathcal{S}(\cdot)$ denotes the SSIM metric, and $\mathbf{M}$ denotes the visible region mask.

For semantic-level fidelity, we use LPIPS~\cite{zhang2018unreasonable} to measure the perceptual distance $lpips$ between the generated video and the source video.

For video quality $q$, we calculate the image quality~\cite{huang2024vbench} of each frame and then take the average over the entire sequence.
We then normalize these metrics and compute the composite reward for the sample as:
\begin{equation}
    r = \mathbf{W} \left[(1-lpips), \frac{p}{p_{th}}, \frac{s}{s_{th}}, q\right]^T,
\end{equation}
where $\mathbf{W} \in \{0, 1\}^{1 \times 4}$ is the weight vector for the consistency-quality reward function, $p_{th}$ and $s_{th}$ are the hyperparameters.
Finally, we use Eq.~\eqref{eq:advantage} to calculate the relative advantage of the sample.

\begin{table}[t]
  \centering
  \caption{Quantitative results of different ADEVIS accumulation methods in kubric-4D at accumulated step=3.}
    \scalebox{0.7}{
    \begin{tabular}{cccccc}
    \toprule
    Settings & PSNR  & SSIM  & LPIPS & PSNR-Nocc & SSIM-Nocc \\
    \midrule
    Baseline w/o LoRA & 9.93  & 0.2753 & 0.6551 & 15.38 & 0.9067 \\
    \midrule
    RFA   & 10.35 & 0.2733 & 0.6240 & 14.81 & 0.8772 \\
    RVA   & 10.83 & 0.2927 & 0.6987 & 12.89 & 0.8973 \\
    ORA-cc & 11.59 & 0.3172 & 0.6083 & 15.62 & 0.9109 \\
    ORA-dt & 10.70 & 0.2815 & 0.6236 & 15.86 & 0.9153 \\
    \rowcolor[rgb]{ .851,  .851,  .851} BTA & 12.62 & 0.3636 & 0.5946 & 16.96 & 0.9279 \\
    \midrule
    Baseline w LoRA & 12.47 & 0.3904 & 0.5264 & 19.02 & 0.9461 \\
    \midrule
    RFA   & 12.93 & 0.3966 & 0.5173 & 22.66 & 0.9532 \\
    RVA   & 13.05 & 0.3782 & 0.6664 & 16.22 & 0.9196 \\
    ORA-cc & 13.51 & 0.4162 & 0.5148 & 22.63 & 0.9553 \\
    ORA-dt & 12.79 & 0.3987 & 0.5189 & 22.45 & 0.9542 \\
    \rowcolor[rgb]{ .851,  .851,  .851} BTA & 14.49 & 0.4519 & 0.4831 & 24.05 & 0.9606 \\
    \bottomrule
    \end{tabular}%
  }
  \label{tab:abl_imts}%
\end{table}%

\begin{table*}[htbp]
  \centering
  \caption{Comparison results between DEVIS-GRPO and Vanilla-GRPO.}
    \begin{tabular}{c|ccccc|cc|c}
    \toprule
    \multirow{2}[4]{*}{Method} & \multicolumn{5}{c|}{Consistency  Metrics} & \multicolumn{2}{c|}{Diversity} & \multirow{2}[4]{*}{Time} \\
\cmidrule{2-8}          & PSNR  & SSIM  & LPIPS & PSNR-Nocc & SSIM-Nocc & CLIP  & LPIPS &  \\
    \midrule
    Vanilla-GRPO & 12.71 & 0.3957 & 0.5698 & 18.48 & 0.9098 & 0.1082 & 0.2189 & 1628 \\
    DEVIS-GRPO & 13.59 & 0.4360  & 0.5289 & 20.17 & 0.9524 & 0.1432 & 0.2677 & 2162 \\
    \bottomrule
    \end{tabular}%
  \label{tab:ablation_grpo}%
\end{table*}%

\begin{figure}[!t]
    \centering
    \includegraphics[width=0.45\textwidth]{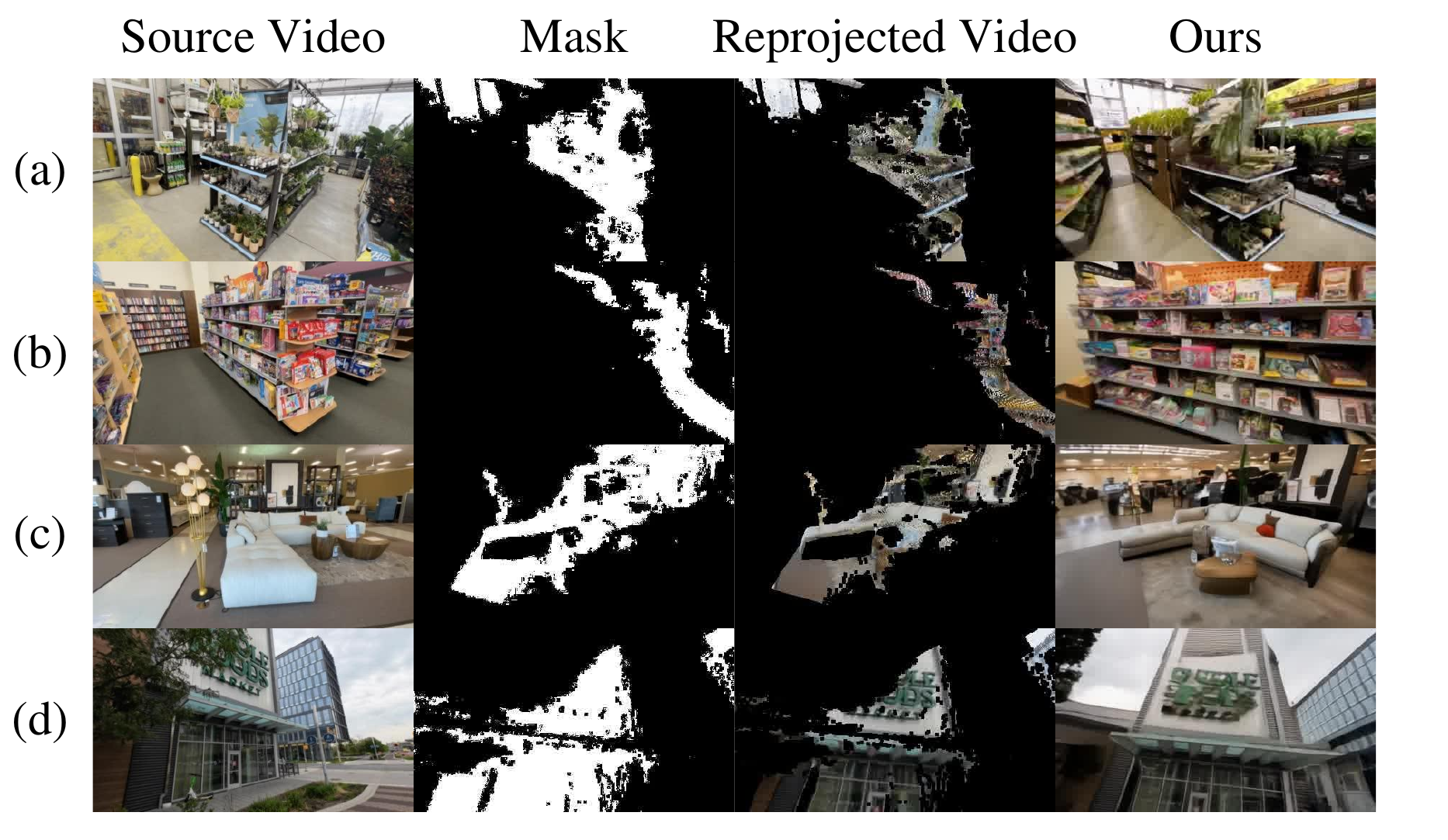}
    \caption{The more visual results on DL3DV.}
    \label{fig:vis_Dl3DV}
\end{figure}

\section{Experiments}
\label{sec:experiments}
\subsection{Experiments Settings}

\paragraph{Implementation Details.}
We add a LoRA~\cite{hu2022lora} layer based on TrajectoryCrafter as our baseline model.
We employ the AdamW optimizer, first fine-tuning on Kubric-4D for small-view video generation for better domain transferability with a learning rate of $1 \times 10^{-4}$, then optimizing with DEVIS-GRPO at a learning rate of $2 \times 10^{-6}$.
Since sampling videos requires considerable time, we train multiple steps per sampling. Specifically, we sample 32 groups at a time, each containing 16 samples, then train for 10 epochs before proceeding to the next sampling round. This process was repeated for a total of 50 rounds.
Training takes 4 days on 8 Ascend 910B NPUs (64GB).

\paragraph{Evaluation Metrics.} 
In multi-view video generation, maintaining consistency between the generated video and source video is crucial. 
Therefore, we employ PSNR, SSIM, and LPIPS~\cite{zhang2018unreasonable} to assess global consistency across the entire video, while using PSNR-Nocc and SSIM-Nocc to measure regional consistency in non-occlusion areas under geometric reprojection.
Additionally, we utilize ``Temporal Flicker" (TF), ``Motion Smooth" (MS), ``Aesthetic Quality" (AQ), and ``Imaging Quality" (IQ) from VBench~\cite{huang2024vbench} to assess the smoothness and quality of the generated video.

\paragraph{Evaluation Datasets.}
We validate the effectiveness of our method on the Kubric-4D, iPhone and DL3DV datasets.
For Kubric-4D~\cite{van2024generative}, we randomly generate 100 camera trajectories with large-angle camera motions ($|\Delta \mathbf{T}| > 120^\circ$) from the test set, and render the corresponding source videos, target videos, and ground-truth depth maps for evaluation.
For iPhone~\cite{gao2022monocular}, we used all five extreme-view video pairs featuring both ground truth video and depth annotations.
Since DL3DV~\cite{ling2024dl3dv} contains static scenes with continuous camera trajectories, we adapt it by using the first frame as the source and frames 10 to 60 as the target sequence, with DepthSplat~\cite{xu2025depthsplat} for more accurate depth estimation.

\subsection{Comparison with State-of-the-Art Methods}

\paragraph{Comparison Methods.} 
We select three representative baselines capable of generating large-angle camera motions for comparison with DEVIS-GRPO: TrajectoryCrafter~\cite{yu2025trajectorycrafter}, EX-4D~\cite{hu2025ex}, and GCD~\cite{van2024generative}. The first two employ explicit reprojection modeling, while the latter adopts an implicit camera embedding modeling.

\paragraph{Quantitative Results.} 
Tab.~\ref{tab:quantitative} presents quantitative comparisons between DEVIS-GRPO and other baselines. 
Our DEVIS-GRPO achieves outstanding performance on consistency metrics.
Most notably, on Kubric-4D's SSIM, LPIPS, PSNR-Nocc, and SSIM-Nocc metrics, our method outperforms the second-best method by 22.6\%, 6\%, 21.6\%, and 7.3\%, respectively. 
This demonstrates that DEVIS-GRPO maintains outstanding global and visible region consistency at both the pixel and semantic levels.

Our approach obtains a lower image quality (IQ) score compared to alternatives. 
This stems from the fact that competing approaches often prioritize generating highly realistic-looking videos, even at the expense of deviating from the ground-truth reference. Please refer to Fig.~\ref{fig:QualitativeKubric} for visualization. 
In contrast, our approach focuses on faithful large-view synthesis.

\paragraph{Qualitative Results.}
Fig.~\ref{fig:QualitativeKubric} and Fig.~\ref{fig:QualitativeIphone} present a visual quality comparison between our method and other methods. 
We observe that while GCD maintains style and background consistency with the source video, it produces blurred details and fails to accurately follow the specified camera trajectory. For example, in Fig.~\ref{fig:QualitativeKubric} column (c), GCD does not follow the camera's movement.
TrajectoryCrafter and EX-4D successfully follow the camera trajectory but exhibit noticeable foreground-background inconsistencies compared to the reprojected video and source video, while also introducing undesired modifications in visible regions. 
We also present visualization results on the real-world dataset DL3DV in Fig.~\ref{fig:vis_Dl3DV}.
In contrast, our DEVIS-GRPO demonstrates superior performance in background-foreground consistency, camera trajectory adherence, and overall visual quality.

\subsection{Ablation Studies}

\paragraph{Ablation on Accumulative Methods.}
Tab.~\ref{tab:abl_imts} presents the effectiveness of different accumulation methods on baseline. 
``w/o LoRA" denotes the model without fine-tuning on Kubric-4D, ``w LoRA" refers to the model fine-tuned.
The results demonstrate that ``BTA" outperforms other accumulation methods and was employed for our GRPO training.

\paragraph{Comparison bettwen DEVIS-GRPO and Vanilla-GRPO.} 

\begin{table}[t]
  \centering
  \caption{The ablation study of different training paradigms. “SFT all,” “SFT best,” and “SFT weight” refer to treating all generated samples equally, training only on the sample with the highest reward score, and using the reward score as the loss weight, respectively.}
      \scalebox{0.75}{
    \begin{tabular}{cccccc}
    \toprule
    Methods & PSNR  & SSIM  & LPIPS & PSNR-Nocc & SSIM-Nocc \\
    \midrule
    STF-all & 12.27 & 0.3892 & 0.5198 & 19.01 & 0.9421 \\
    STF-best & 12.54 & 0.4065 & 0.5268 & 19.15 & 0.9483 \\
    STF-weight & 12.51 & 0.4066 & 0.5287 & 19.06 & 0.9479 \\
    DPO   & 12.25 & 0.3764 & 0.5388 & 18.81 & 0.9433 \\
    DEVIS-GRPO & 13.59 & 0.4360 & 0.5289 & 20.17 & 0.9524 \\
    \bottomrule
    \end{tabular}%
    }
  \label{tab:ablation_rl}%
\end{table}%

To demonstrate the superiority of DEVIS-GRPO over Vanilla-GRPO, we compared their performance metrics and runtime per iteration. As shown in Tab.~\ref{tab:ablation_grpo}, although DEVIS-GRPO incurs a slight increase in runtime, it achieves better consistency metrics and generates more diverse samples.

\paragraph{Ablation on Training Paradigms.} 
In addition, to demonstrate the effectiveness of RL training, we further compared our approach with SFT and DPO.
As shown in Tab.~\ref{tab:ablation_rl}, DEVIS-GRPO still achieves the best performance, benefiting from GRPO’s use of group-relative advantages to distribute rewards across the entire group, as well as its iterative online policy improvement through sampling.

\section{Conclusion}
This paper proposes DEVIS-GRPO, pioneering the integration of GRPO into camera-controlled video generation to achieve extreme viewpoints generation.
The key of DEVIS-GRPO lies ADEVIS, which decomposes large-angle camera motions into a series of manageable small-angle steps, enabling baseline to accumulate positive views. ADEVIS eliminates the need for expensive paired large-angle video data and enhances sampling diversity.
To explore ADEVIS, we propose four different accumulation strategies to fuse prior content from preceding steps.
Additionally, we design a consistency-quality reward function to evaluate the relative advantages.
Experiments demonstrate the superiority of DEVIS-GRPO under dynamic extreme view.

\bibliography{aaai2026}


\end{document}